\documentclass[10pt,twocolumn,letterpaper]{article}
\usepackage{wacv}
\usepackage{times}
\usepackage{epsfig}
\usepackage{graphicx}
\usepackage{amsmath}
\usepackage{amssymb}
\usepackage{arydshln}
\usepackage{colortbl}
\usepackage{color}
\usepackage{float}
\usepackage{caption}
\usepackage{float}
\usepackage{cite}
\makeatletter

\wacvfinalcopy % *** Uncomment this line for the final submission

\def\wacvPaperID{129} % *** Enter the wacv Paper ID here

\ifwacvfinal\fi
\begin{document}

%%%%%%%%% TITLE
%\title{On the Utilization of Multi-component Image Translation for Domain Generalization}

\title{Multi-component Image Translation for Deep Domain Generalization}

% Authors at the same institution
\author{Mohammad Mahfujur Rahman
\hspace{0.5cm} Clinton Fookes
\hspace{0.5cm} Mahsa Baktashmotlagh  \hspace{0.5cm} Sridha Sridharan \\
Image and Video Laboratory, Queensland University of Technology (QUT), Brisbane, QLD, Australia\\
{\tt\small Email: \{m27.rahman, c.fookes,  m.baktashmotlagh, s.sridharan\}@qut.edu.au}
}
% Authors at different institutions
%\author{First Author \\
%\and
%Second Author \\

\maketitle
\ifwacvfinal\thispagestyle{empty}\fi

%%%%%%%%% ABSTRACT
\begin{abstract}
  % Domain adaption (DA) and domain generalization (DG) are two closely related methods which are both concerned with the task of assigning labels to an unlabeled data set. The only dissimilarity between these approaches is the accessibility  of the target samples during the training phase. DA can access the target data during training, while the target data is totally unseen during training in DG. The task of DG is challenging as we have no earlier knowledge of the target samples. The aim of DG is to acquire knowledge from the multiple source domains available to obtain a domain-independent model that can be used for a particular task, such as classification, to an unseen domain. If DA methods are applied directly to DG by a simple exclusion of the target data from training, poor performance will result for a given task. In this paper, we propose a novel deep domain generalization architecture utilizing synthetic data generated by a Generative Adversarial Network (GAN). The discrepancy between the generated images and synthetic images is minimized using existing domain discrepancy metrics such as maximum mean discrepancy or correlation alignment. Furthermore, we introduce a protocol for applying DA methods to a DG scenario by excluding the target data from the training phase, splitting the source data to training and validation parts,  and treating the validation data as target data for DA. We conduct extensive experiments on four cross-domain benchmark datasets. Experimental results signify our proposed model outperforms the current state-of-the-art methods for DG.

 Domain adaption (DA) and domain generalization (DG) are two closely related methods which are both concerned with the task of assigning labels to an unlabeled data set. The only dissimilarity between these approaches is that DA can access the target data during the training phase, while the target data is totally unseen during the training phase in DG. The task of DG is challenging as we have no earlier knowledge of the target samples. If DA methods are applied directly to DG by a simple exclusion of the target data from training, poor performance will result for a given task. In this paper, we tackle the domain generalization challenge in two ways. In our first approach, we propose a novel deep domain generalization architecture utilizing synthetic data generated by a Generative Adversarial Network (GAN). The discrepancy between the generated images and synthetic images is minimized using existing domain discrepancy metrics such as maximum mean discrepancy or correlation alignment. In our second approach, we introduce a protocol for applying DA methods to a DG scenario by excluding the target data from the training phase, splitting the source data to training and validation parts,  and treating the validation data as target data for DA. We conduct extensive experiments on four cross-domain benchmark datasets. Experimental results signify our proposed model outperforms the current state-of-the-art methods for DG.
   
\end{abstract}

\section{Introduction}

The success of Deep Neural Networks (DNNs) or Deep Learning (DL) largely depends on the availability of large sets of labeled data. When the training and test (source and target) images come from different distributions (domain shift), DL cannot perform well. Domain adaption (DA) and domain generalization (DG) methods have been proposed to address the poor performance due to such domain shift. DA and DG are similar frameworks, but the key difference between DA and DG is the availability of the target data in training phase. DA methods access the target data during training while DG approaches do not have access to the target data.

%In the presence of large sets of labeled data, Deep Learning (DL) has accomplished extraordinary triumphs in the avenue of computer vision, particularly in object classification and recognition tasks. However, DL cannot always perform well when the training and testing (source and target) images come from different distributions, in the presence of domain shift between training, or testing images and in the absence of enough labeled data (target data). DA and DG methods have been proposed to address the poor performance due to such domain shift. Domain adaptation and domain generalization are similar frameworks, but the key difference between DA and DG is the availability of target data during training. DA methods access target data during training while DG approaches do not have access to the target data. 

%such as object recognition \cite{NIPS2012_4824}, object detection \cite{Girshick:2014:RFH:2679600.2679851}, person re-identification \cite{8354250}, speech recognition \cite{6639346} and machine translation \cite{DBLP:journals/corr/SutskeverVL14}.

DA is an attractive area of research in computer vision and pattern recognition fields because it handles the task properly with limited target samples. Domain adaptation can be classified into three groups: Unsupervised Domain Adaptation (UDA), Semi-Supervised Domain Adaptation (SSDA) and Supervised Domain Adaptation (SDA). UDA does not require any labeled target data whereas SDA needs all labeled target samples. Unsupervised deep domain adaptation (UDDA) methods require large sets of target data in order to be more successful in producing a desired output.

To reduce the discrepancy between the source and target data, traditional domain adaptation methods consider seeking domain-invariant information of the data, adapting classifiers, or both. In the conventional domain adaptation task, we still access the target data in training phase. However, in real world scenarios, the target data are out of reach in the training phase. Consequently, ordinary domain adaptation methods do not perform well without having target data in the training phase.

When target data is inaccessible, the issue of DG arises. Without the target data, DG makes full use of and derives benefit from all available source domains. It then tries to acquire a domain independent model by merging data sources which can be used for unknown target data. For an example, we have the images from ImageNet \cite{5206848} and Caltech-256 \cite{griffinHolubPerona} datasets and we want to classify images in the LabelMe dataset \cite{Russell2008}. We have source data from multiple sources for training without having any knowledge about the target data. That is, the target data is completely unseen during the training phase. DG models utilize the captured information from various source domains and generalize to unseen target data, which is used only in the test phase. Although DA and DG frameworks are very close and both have similar goals (producing a strong classifier on target data), the existing DA techniques do not perform well when directly applied in DG. 

In this paper, we propose a novel deep domain generalization framework by utilizing synthetic data  that are generated by a GAN during the training stage where the discrepancy between the real and synthetic data is minimized using the existing domain adaptation metrics such as maximum mean discrepancy or correlation alignment. The motivation behind this research is that if we can generate different images from a single image whose styles are similar to the available source domains, the framework will learn a more agnostic model that can be applied for unseen data. Our approach takes advantage of the natural image correspondence built by CycleGAN \cite{CycleGAN2017} and ComboGAN \cite{DBLP:journals/corr/abs-1712-06909}. The contributions of this paper are two-fold: 

\begin{itemize}

\item We implement a novel deep domain generalization framework utilizing synthetic data that is generated by a GAN. We generate images from one source domain into the distributions of other available source domains which are used in the training phase as validation data. The discrepancy between the real data and synthetic data is decreased using existing domain discrepancy metrics in DG settings.

\item We introduce a protocol for applying DA methods on DG scenarios where the source datasets are split into training and validation sets, and the validation data acts as target data for DG.

%70\% data is fed into one stream of CNN while the rest of the data (30\%) is fed into another stream of CNN during training, and the discrepancy between these splitting data are minimized using the existing domain adaptation metrics.

%\item The third contribution of this work is to release synthetic datasets for domain generalization.
\end{itemize}

We conduct extensive experiments to evaluate the image classification accuracy of our proposed method across a large set of alternatives in DG settings. Our method performs significantly better than previous state-of-the-art approaches across a range of visual image classification tasks with domain mismatch between the source and target data.

%\subsection{Organization of the Paper}

\section{Related Research}
This section reviews existing research on DA and DG, especially in the avenue of object classification. %Domain adaptation and domain generalization have recently pulled in extraordinary enthusiasm for deep learning. We address a recent literature review of domain adaptation and domain generalization that is sorted out into two sections: domain adaptation and domain generalization.

\subsection{Domain Adaptation}

To address the problem of domain shift, many DA methods \cite{Zadrozny:2004:LEC:1015330.1015425, NIPS2007_3248, 5640675, DBLP:conf/cvpr/GongSSG12,  DBLP:journals/corr/TzengHZSD14, DBLP:journals/corr/TzengHDS15, DBLP:journals/corr/Long015, DBLP:journals/corr/Long0J16, DBLP:journals/corr/Long0J16a, venkateswara2017Deep} have been introduced in recent years. All the DA techniques can be divided into two main categories: Conventional Domain Adaptation methods \cite{Zadrozny:2004:LEC:1015330.1015425, NIPS2007_3248, 5640675, DBLP:conf/cvpr/GongSSG12} and Deep Domain Adaptation methods \cite{DBLP:journals/corr/TzengHZSD14, DBLP:journals/corr/TzengHDS15, DBLP:journals/corr/Long015, DBLP:journals/corr/Long0J16, DBLP:journals/corr/Long0J16a, venkateswara2017Deep}. The conventional domain adaptation methods have been developed into two phases, the features are extracted in the first phase and these extracted features are used to train the classifier in the second phase. However, the performance of these DA methods is not satisfactory. 

%There have been many DA methods proposed in recent years to solve the problem of domain shift. All the methods can be categorized into two main categories, Conventional Domain Adaptation and Deep Domain Adaptation methods. The conventional domain adaptation methods develop their model into two stages, feature extraction and classification. In the first phase, these methods extract features and in the second phase, a classifier is trained to classify the objects. However, the performance of these DA methods is not satisfactory. 

%Obtaining the features using a deep neural network even without adaptation techniques outperforms conventional DA methods by a large margin. The image classification accuracy obtained with Deep Convolutional Activation Features (DeCAF) \cite{Donahue:2014:DDC:3044805.3044879} even without using any adaptation algorithm is remarkably better than any conventional domain adaptation methods due to the capacity of a Deep Neural Network (DNN) to extract more robust features using nonlinear function. As a result, deep neural network based domain adaptation methods are gaining strong popularity day by day.

Obtaining the features using a deep neural network even without adaptation techniques outperforms conventional DA methods by a large margin. The image classification accuracy obtained with Deep Convolutional Activation Features (DeCAF) \cite{Donahue:2014:DDC:3044805.3044879} even without using any adaptation algorithm is remarkably better than any conventional domain adaptation methods \cite{Zadrozny:2004:LEC:1015330.1015425, NIPS2007_3248, 5640675, DBLP:conf/cvpr/GongSSG12} due to the capacity of a DNN to extract more robust features using nonlinear function. Inspired by the success of DL, researchers widely adopted DL based DA techniques.

Maximum Mean Discrepancy (MMD) is a well known metric for comparing the discrepancy between the source and target data. Eric Tzeng et al. \cite{DBLP:journals/corr/TzengHZSD14} introduced the Deep Domain Confusion (DDC) domain adaptation method where the discrepancy is reduced by introducing a confusion layer. In \cite{DBLP:journals/corr/TzengHDS15}, the previous work \cite{DBLP:journals/corr/TzengHZSD14} is improved by adopting a soft label distribution matching loss which is used to reduce the discrepancy between the source and target domains. Long et al. proposed the Domain Adaptation Network (DAN) \cite{DBLP:journals/corr/Long015} that introduced the sum of MMDs defined between several layers which are used to mitigate the domain discrepancy problem. This idea was further boosted by the Joint Adaptation Networks \cite{DBLP:journals/corr/Long0J16a} and Residual Transfer Networks \cite{DBLP:journals/corr/Long0J16}. Hemanth et al. \cite{venkateswara2017Deep} proposed a new Deep Hasing Network for UDA where hash codes are used to address the DA problem.

Correlation Alignment is another popular metric for feature adaptation where the covariances or second order statistics of the source and target data are aligned. In \cite{coral,dcoral,DBLP:journals/corr/MorerioM17}, UDDA approaches have been proposed where the discrepancy between the source and target data is reduced by aligning the second order statistics of the source and target data. The idea of \cite{coral,dcoral,DBLP:journals/corr/MorerioM17} is similar to Deep Domain Confusion (DDC) \cite{DBLP:journals/corr/TzengHZSD14} and Deep Adaptation Network (DAN) \cite{DBLP:journals/corr/Long015} except that instead of MMD, they adopted CORAL loss to minimize the discrepancy. CORAL loss is modified in \cite{DBLP:journals/corr/KoniuszTP16} by introducing a Riemannian Metric which performs better than \cite{dcoral,DBLP:journals/corr/MorerioM17}. The aforementioned methods utilized two streams of Convolutional Neural Networks (CNN) where the source and target networks are fused at the classifier level. Domain-Adversarial Neural Networks (DANN) \cite{Ganin:2016:DTN:2946645.2946704} introduced a deep domain adaptation approach by integrating a gradient reversal layer into the traditional architecture.

Current DA approaches still suffer from DG problems when the target data is unavailable during training. In this paper, we explore how the DA approaches can be applied more efficiently on DG scenarios to improve their generalization capability. 

%If DA methods are applied directly to DG by a simple exclusion of the target data from training, poor performance will result for a given task

\subsection{Domain Generalization}
In recent research, DG is a less explored issue than DA. The aim of DG is to acquire knowledge from the multiple source domains available to obtain a domain-independent model that can be used for a particular task, such as classification, to an unseen domain. Blanchard et al. \cite{NIPS2011_4312} first introduced an augmented Support Vector Machine (SVM) based model that solved automatic gating of flow cytometry by encoding empirical marginal distributions into the kernel. In \cite{Muandet:2013:DGV:3042817.3042820}, a DG model was proposed which learned a domain invariant transformation by reducing the discrepancy among available source domains. Xu et al. \cite{10.1007/978-3-319-10578-9_41} proposed a domain generalization method that worked on unseen domains by using low-rank structure from various latent source domains. Ghifary et al. \cite{DBLP:conf/iccv/GhifaryKZB15} proposed an auto-encoder based technique to extract domain-invariant information through multi-task learning. These DG methods aggregate all the samples from training datasets to learn a shared invariant representation which is then used for the unseen target domain.

In \cite{Khosla:2012:UDD:2402940.2402953}, a domain generalization technique was introduced based on a multi-task max-margin classifier that regulates the weights of the classifier to improve the performance on unseen datasets which was inaccessible in the training phase. Fang et al. \cite{6751316} proposed a domain generalization method using Unbiased Metric
Learning (UML). It makes a less biased distance metric that provides better object classification accuracy for an unknown target dataset. In \cite{10.1007/978-3-319-10578-9_41}, a domain generalization approach was proposed that combined the score of the classifiers. For multi-view domain generalization, \cite{7410834,7733141} proposed another DG approach where multiple types of features of the source samples were used to learn a robust classifier. All the above methods exploit all the information from the training data to train a classifier or adjust weights. 

The above mentioned DG methods use shallow models, hence the performance still needs to improve. Moreover, these shallow models extract the features first, which are then used to train the classifier. As a result, these are not end-to-end methods. Motiian et al. \cite{motiian2017CCSA} proposed supervised deep domain adaptation and generalization using contrastive loss to align the source data and target data. In this method, the source domains are aggregated and the learned model from source data is used for unseen target data. Li et al. \cite{8237853} proposed another DG method based on a low-rank parameterized CNN. A deep domain generalization
architecture with a structured low-rank constraint to mitigate the domain generalization issue was proposed in \cite{8053784} where consistent information across multiple related source domains were captured.

%Although many DA techniques based on deep architecture are proposed in recent years, very few DG approaches based on deep architecture are introduced. In this paper, we explore GAN as well as deep architecture to address the DG challenges. We propose a deep domain generalization framework using synthetic data generated by a GAN where the styles are transfered from one domain into another domain.

Although many DA techniques based on deep architectures are proposed in recent years, very few DG approaches based on deep architecture are introduced. In this paper, we explore GAN along with a deep architecture to address the DG challenges. We propose a deep domain generalization framework using synthetic data generated by a GAN where the styles are transferred from one domain into another domain.

\section{Methodology}

In this section, we will introduce our proposed Deep Domain Generalization approach by using synthetic data during training to produce an agnostic classifier that can be applied for unseen target data. We build a model to generate images with the distributions of one source domain into other available source domains. We follow the definitions and notation of \cite{DBLP:journals/pami/GhifaryBKZ17,7078994,DBLP:series/acvpr/978-3-319-58347-1}. Let, $X$ and $Y$ stand for the input (data) and the corresponding output (label) random variables, and $P(X,Y)$ represents the joint probability distribution of the input and output. We can define a domain as the probability distribution $P(X,Y)$ on $X \times Y$. We formalize domain adaptation and domain generalization as follows:

\begin{figure*}
\begin{center}
\includegraphics[width=0.61\linewidth]{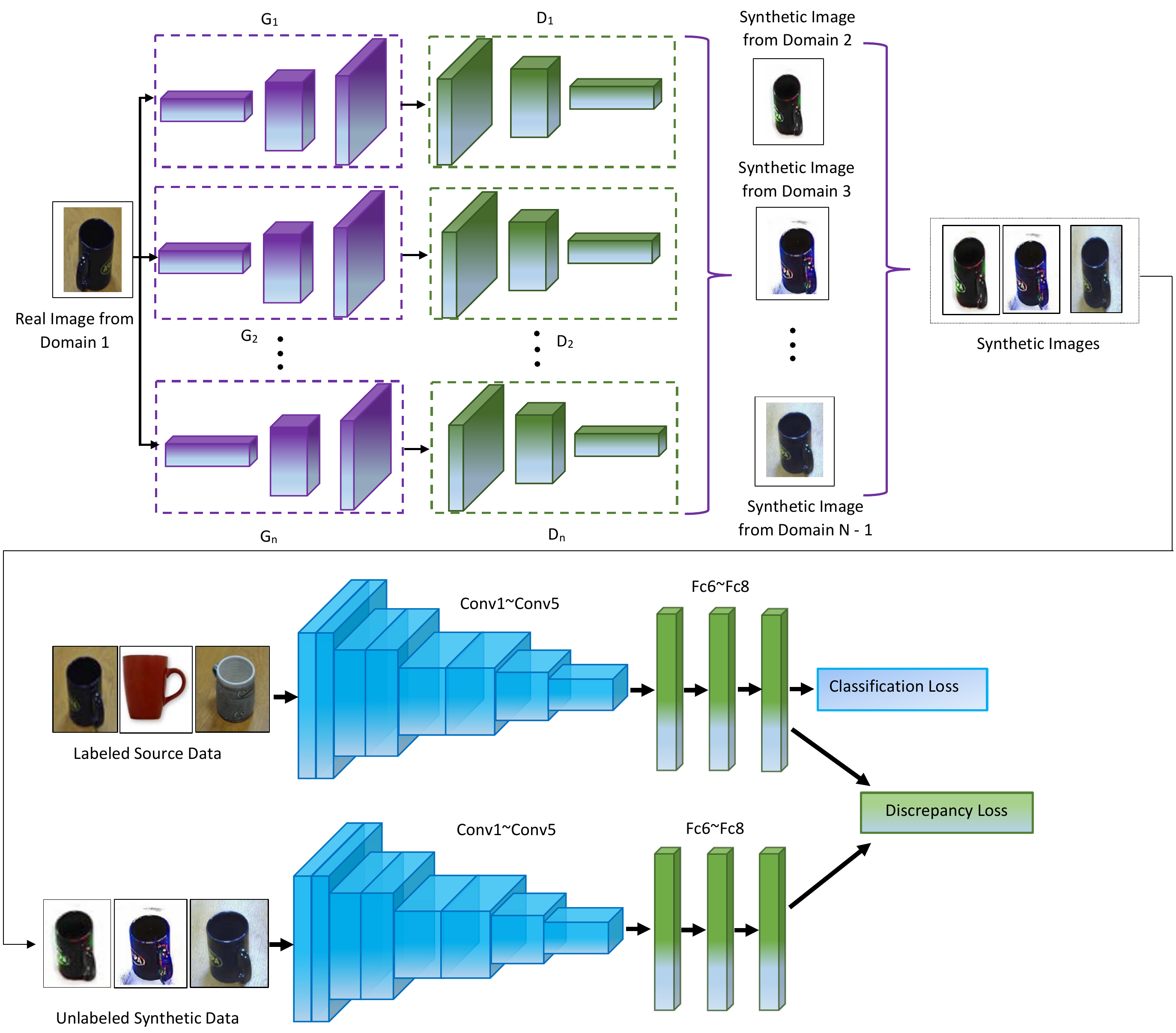}
\end{center}
   \caption{Our proposed methodology of Deep Domain Generalization. At first, synthetic images are generated using ComboGAN \cite{DBLP:journals/corr/abs-1712-06909} which are used during the training phase and a discrepancy loss is used to mitigate the distribution discrepancy between the real source domain images and generated synthetic images. For an example, in Office-Caltech dataset \cite{DBLP:conf/cvpr/GongSSG12}, 4 different domains are available: Amazon, Webcam, DSLR and Caltech. When Caltech domain is unseen, we generate the images from the other three domains: Amazon, Webcam and DSLR. These generated images are used to train the network while the images from Caltech domain are classified during the test phase.}
\label{fig:architecture}
\end{figure*}

\subsubsection*{Domain Adaptation}  

Let us consider that the source domain data samples are $S_D=\{X_i^s\}$ with available labels $L_s=\{Y_i\}$ and the target data samples are $T_D = \{X_i^t\}$ without labels. The data distribution of the source and target data are different, i.e., $P_s(X_i^s,Y_s)$ $\neq$ $P_t(X_i^t,Y_t)$ where $Y_t$ is the label of target samples. The task of domain adaptation is to learn a classifying function $f: X \rightarrow Y$ able to classify $\{X_i^t\}$ to the corresponding $\{Y_i^t\}$ given $S_D$ and $T_D$ during training as the input. 

\subsubsection*{Domain Generalization}
Let us consider that we have a set of $n$ number of source domains such as $\Delta$ = $S_D^1$; $S_D^2$; $S_D^3$;\dots;$S_D^n$ and target domain $T_D^n$ where $T_D^n$ $\notin$ $\Delta$. The aim of domain generalization is to gain a classifying function $f: X \rightarrow Y$ able to classify $\{X_i^t\}$ to the corresponding $\{Y_i^t\}$ given $S_D^1$; $S_D^2$; $S_D^3$;\dots;$S_D^n$ during training as the input, but $T_D^n$ is unavailable during training.

 %\subsubsection{Image-to-Image Translation}

\subsection{Our Approach}

Generative Adversarial Networks (GANs) have accomplished great outcomes in different applications, for example, image generation \cite{NIPS2014_5423}, text2image \cite{Reed:2016:GAT:3045390.3045503}, image-to-image translation \cite{DBLP:conf/cvpr/IsolaZZE17, CycleGAN2017},  person re-identification \cite{8354250} and image editing \cite{Perarnau2016}. The strong success of GANs is influenced by the concept of an adversarial loss that drives the generated images to be identical to the real images. Zhu et al. \cite{bicycleGAN} proposed a conditional multimodal image-to-image translation method where the training procedure requires paired data which is a supervised solution of image translation from one domain to other domains. As we do not have paired data on domain adaptation or generalization settings, we do not consider this method for synthetic image generation. Zhu et al. \cite{CycleGAN2017} proposed CycleGAN that learned a mapping between an input image and an output image using both adversarial and cycle consistency loss where unpaired data were used for image generation.  Anoosheh et al. \cite{DBLP:journals/corr/abs-1712-06909} modified CycleGAN by reducing the number of generators and discriminators while there are more than two domains. Choi et al. \cite{starGAN} proposed Stargan for image translation from one domain to another domain. Huang et al. \cite{MUNIT} proposed another multi-modal image translation technique known as MUNIT which can translate image from one domain to other domains. However, the experiments so far have been tailored by \cite{starGAN,MUNIT} to merely two domains at a time whereas ComboGAN\cite{DBLP:journals/corr/abs-1712-06909} is capable to translate images at a time when more than two domains exist in the dataset. Our approach takes advantage of the natural image correspondence built by \cite{DBLP:journals/corr/abs-1712-06909}. Figure \ref{fig:architecture} shows the overall architecture of our proposed method for DG based on synthetic image. We use ComboGAN \cite{DBLP:journals/corr/abs-1712-06909} to translate the samples in the source domain $D_s^n$ to those other domains. Having $4$ domains of Office-Home dataset \cite{venkateswara2017Deep}, we take the image from one domain i.e., $Art$ and translate it to 4 different image domains which are
similar to $Art$, $Clipart$, $Product$ and $Real-world$ domains.

GANs comprise of a generator $G$ and a discriminator $D$. The input of the generator is random numbers and it generates an image. This generated image is fed into the discriminator with a stream of images taken from real datasets and the discriminator returns the probabilities representing the probability of a real or fake prediction. The training process is called adversarial training. Training itself is executed in two alternating steps; first the discriminator $D$ is trained to distinguish between one or more pairs of real and generated samples, and the generator is trained to fool the discriminator with generated samples. At beginning, the discriminator should be too powerful so that it can identify the real and fake images. After that the generator would be more powerful than the discriminator so that the discriminator cannot distinguish the real and generated images. The training procedure of GAN is a min-max game between a generator and a discriminator. The optimization problem of GAN can be expressed as,

\begin{align}
\underset{G}{min} \:  \underset{D}{max} \: E_{x}[log D(x)] + E_z[log(1 - D(G(z))].
\end{align}

%The aim of our model is to find the mapping functions among $n$ number of domains $S_D^1$; $S_D^2$; $S_D^3$;\dots;$S_D^n$ where the training samples are ${X_1^s}$, ${X_2^s}$, ${X_3^s}$,\dots,${X_n^s}$. Our models includes $n$ number of mappings $G_1$: $S_D^1$ $\rightarrow$ $S_D^2$; $G_2$: $S_D^1$$\rightarrow$ $S_D^3$;\dots; $G_n$: $S_D^1$$\rightarrow$ $S_D^n$. We use $n$ number of adversarial discriminators $D_x$, $D_y$,\dots,$D_n$.

Let the available source domains be $S_D^1$; $S_D^2$; $S_D^3$;\dots;$S_D^n$. The training samples from the domains  are $\{X_i^1\}, \{X_i^2\}, \{X_i^3\},\dots \{X_i^n\} $ respectively. The corresponding data distribution of the domains are $ P_s^1, P_s^2, P_s^3,\dots,P_s^n.$ The aim of our model is to learn mapping functions among the available source domains $S_D^1$; $S_D^2$; $S_D^3$;\dots;$S_D^n$. Our model includes $n$ mappings $G_1$: $S_D^1$ $\rightarrow$ $S_D^2$; $G_2$: $S_D^1$$\rightarrow$ $S_D^3$;\dots; $G_n$: $S_D^1$$\rightarrow$ $S_D^n$. In our model, we have $n$ number of discriminators where the discriminators distinguish between the real images and translated images. For transferring one domain's images into the style of other domain's images, we use adversarial losses \cite{NIPS2014_5423} and cycle consistency losses \cite{CycleGAN2017}. To compare the distribution of the synthetic images to the distribution in the target domain, adversarial losses are used. On the other hand, to preclude the learned mappings $G_1$, $G_2$, $G_3$ and $G_n$ to be in conflict with each other, cycle consistency losses are used. The adversarial losses, for an example, mapping function  $G_1: S_D^1 \rightarrow S_D^2$ and its discriminator $D_1$ can be expressed as,
%equation of cycle GAN

\begin{align}
L_{GAN}(G_1, D_1, S_D^1, S_D^2) = E_{y \sim P_{S2}} [log D_1(y)]  \nonumber	\\
+ E_{x \sim P_{S1}} [log(1-D_1(G_1(x))],
\end{align}

\noindent where the generator attempts to generate images with the distribution of a new domain ($S_D^2$) and the discriminator $D_1$ tries to differentiate the generated images from the real images.

We need to map an individual input $x_i$ to an output $y_i$, however, adversarial losses cannot ensure that it can map the input to the output properly. In addition to decrease the discrepancy among mapping functions, we use cycle consistency loss, 

\begin{align}
L_{cyc}(G_1, G_2) = E_{x \sim P_{s1}} [ || G_2(G_1(x)) - x||] \nonumber	\\
+ E_{y \sim P_{s2}} [||G_1(G_2(y)) -y||].
\end{align}

To transfer the style of one domain into another domain, for an example, to generate images in the style in domain $S_D^2$ from the image of domain $S_D^1$, the full objective of the GAN is,

\begin{align}
L(G_1, G_2, D_1, D_2) =  L_{GAN}(G_1, D_2, S_D^2, S_D^1) \nonumber \\
+ L_{GAN}(G_2, D_1, S_D^2,S_D^1) \nonumber \\ 
+ \hspace{8ex}{\lambda L_{cyc}(G_1,G_2)}.
\label{eq:3}
\end{align}

\subsection{Domain Generalization using Synthetic Data}

The previous domain generalization method \cite{DBLP:journals/pami/GhifaryBKZ17} divided the source domain datasets into a training set and a test set by random selection from the source datasets. It minimized the discrepancy between these training and test samples during training. In our method, we decrease the discrepancy between the real images and generated images. If there is no domain mismatch between the source and target data, the classifying function $f$ is trained by minimizing the classification loss,
\begin{align}
L_C(f) = E[l(f(X^s),Y],
\end{align}

\noindent where $E[.]$ and $l$ represent mathematical expectation and any loss functions (such as, categorical cross-entropy for multi-class classification) respectively. On the other hand, if the distribution of the source and target data are different, a DA algorithm is used to minimize the  discrepancy between the source and target distributions. As in UDA techniques, there is no labeled data, thus the discrepancy is minimized using the following Equation,

\begin{align}
L_D(g) = d (p(g(X^s)), p(g(X^t))).
\end{align}

The motivation behind Equation (6) is to adjust the distributions of the features within the embedding space, mapped from the source and the target data. Overall, in most of the deep domain adaptation algorithms, the objective is, 

\begin{align}
L_T(f) = L_C + L_D.
\end{align}

We use GAN to generate images from one source domain into the distributions of other available source domains. The discrepancy between the synthetic and real images is minimized by domain discrepancy loss during training. Thus, the learned model can be applied to an unseen domain more effectively. We propose the new loss for training the model as follows,

%In this work, we generate synthetic images which are similar to other domain images, and minimize the discrepancy between the synthetic and real images using domain discrepancy loss during training. Thus, the learned model can be applied to an unseen domain more effectively. We proposed the new loss for training the model,

\begin{align}
L_D(g) = d (p(g(X^s)), p(g(X^G))),
\end{align}

\noindent where $X^s$ are the real images and
$X^G$ are the synthetic images, $d$ can be any domain discrepancy metric such as coral loss or maximum mean discrepancy.

\subsection{Protocol for Applying DA Methods on DG Scenario}
In this section, we will discuss our second approach for domain generalization where we randomly split each source domain data into a training ($70\%$) and a validation set ($30\%$). Most of the DA methods use Siamese based architecture where two stream of CNN is used. This $70\%$ (from all source domain) is fed into one stream of CNN and the other $30\%$ (from all source domain) data is fed into the second stream of CNN during the training phase. The discrepancy is minimized between these data in the training phase. In the test phase, we use unlabeled target data which is completely unseen during the training stage to evaluate our method for domain generalization.

\section{Experiments}

In this section, we conduct substantial experiments to evaluate the proposed method and compare with state-of-the-art UDDA and DG techniques.

\subsection{Datasets}
We evaluate all the methods on four standard domain adaptation and generalization benchmark datasets: Office-31 \cite{Saenko:2010:AVC:1888089.1888106}, Office-Caltech \cite{DBLP:conf/cvpr/GongSSG12}, Office-Home \cite{venkateswara2017Deep} and PACS \cite{8237853}.

%\subsubsection{Office-31}

\textbf{Office-31} \cite{Saenko:2010:AVC:1888089.1888106} is the most prominent benchmark dataset for domain adaptation. It has 3 domains: \textbf{Amazon (A)} domain is formed with the downloaded images from amazon.com, \textbf{DSLR (D)}, containing images captured by Digital SLR camera, and \textbf{Webcam (W)} contains images that captured by web camera with different photo graphical settings. For all experiments, we use labeled source data and unlabeled target data for unsupervised domain generalization where the target data is totally unseen during training. We conduct experiments on 3 transfer tasks ($A, W$ $\rightarrow$ $D$; $A, D$ $\rightarrow$ $W$ and $D, W$ $\rightarrow$ $A$) where two domains are used as source domains and other is as target domain. For an example, the transfer task of $A, W$ $\rightarrow$ $D$, $Amazon(A)$ and $Webcam(W)$ are used as source domains whereas $DSLR(D)$ is used as target domain. We also calculate the average performance of all transfer tasks.

%It contains 4110 images with 31 object categories from an office environment

%\textbf{Office-31} \cite{Saenko:2010:AVC:1888089.1888106} is the most prominent benchmark dataset for domain adaptation in the context of image classification.

\begin{figure}
\begin{center}
\includegraphics[width=0.73\linewidth]{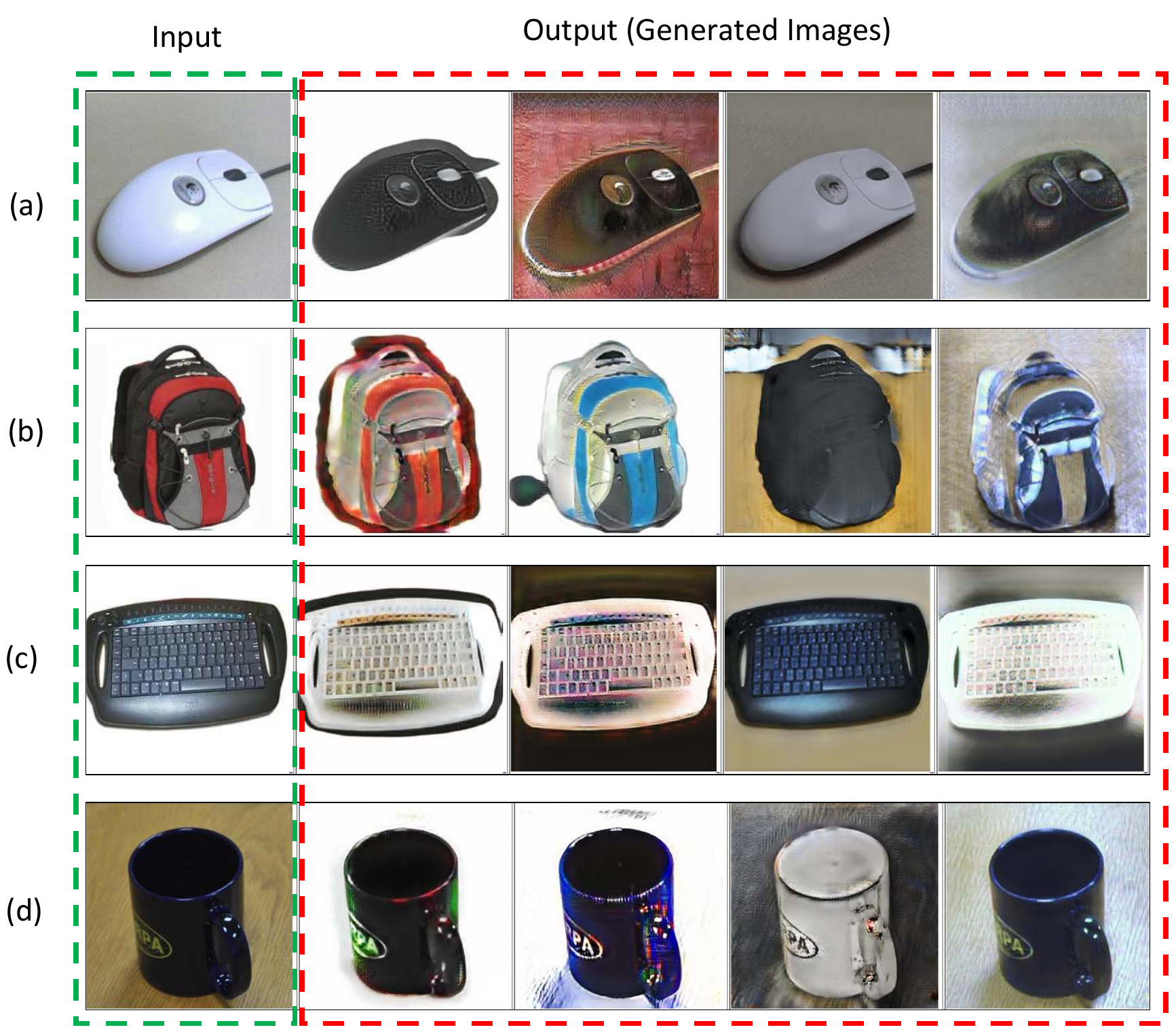}
\end{center}
   \caption{Sample synthetic images from the \textbf{Office-Caltech} dataset \cite{DBLP:conf/cvpr/GongSSG12} which comprises 4 domains. $(a)$ Four images are generated from one image of Webcam domain, $(b)$ four images are generated from one image of Amazon domain, $(c)$ four images are generated from one image of Caltech domain and $(d)$ four images are generated from one image of DSLR domain. }
\label{fig:office}
\end{figure}

%\subsubsection{Office-Caltech}

\textbf{ Office-Caltech} \cite{DBLP:conf/cvpr/GongSSG12} dataset is formed by taking the 10 common classes of two datasets: Office-31 and Caltech-256. It has 4 domains: \textbf{Amazon (A)}, \textbf{Webcam (W)}, \textbf{DSLR (D)} and \textbf{Caltech (C)}. We conduct experiments on 4 transfer tasks where 3 domains or 2 domains are used as source domains ($W, D, C$ $\rightarrow$ $A$; $A, W, D$ $\rightarrow$ $C$; $A, C$ $\rightarrow$ $D, W$ and $D, W$ $\rightarrow$ $A, C$).

%\subsubsection{Office-Home}

\textbf{Office-Home} \cite{venkateswara2017Deep} is a benchmark dataset for domain adaptation which contains 4 domains where each domain consists of 65 categories. The four domains are: \textbf{Art (Ar)}: artistic images in the form of
sketches, paintings, ornamentation, etc.; \textbf{Clipart (Cl)}: collection of clipart images; \textbf{Product (Pr)}: images of objects without a background and \textbf{Real-World (Rw)}: images of objects captured with a regular camera. It contains 15,500 images, with an average of around 70 images per class and a maximum of 99 images in a class. We conduct experiments on 4 transfer tasks where 3 domains are used as source domains ($Pr, Rw, Cl$ $\rightarrow$ $Ar$; $Pr, Rw, Ar$ $\rightarrow$ $Cl$; $Pr, Ar, Cl$ $\rightarrow$ $Rw$ and $Ar, Cl, Rw$ $\rightarrow$ $Pr$).

%\subsubsection{PACS} 
\textbf{PACS} \cite{8237853} is also a recently released benchmark dataset for domain generalization which is created by considering the common classes among Caltech256 , Sketchy, TU-Berlin and Google Images. It has 4 domains, each domain consists of 7 categories. It contains total 9991 images. We conduct experiments on 4 transfer tasks where 3 domains are used as source domains ($P, C, S$ $\rightarrow$ $A$; $P, S, A$ $\rightarrow$ $C$; $S, A, C$ $\rightarrow$ $P$ and $P, C, A$ $\rightarrow$ $S$).

\subsection{Network Architecture}

We adopt the network model for generative architecture as \cite{DBLP:journals/corr/abs-1712-06909,CycleGAN2017} where impressive results for image translation in different domains are shown. The GAN network consists of two convolution layers with two stride-2, several residual blocks and two fractionally strided convolution layers with stride $0.5$. We use 9 blocks for $256$ $\times$ $256$ resolution images. We use $70$ $\times$ $70$ PatchGANs for the discriminator networks. For domain generalization, we use DAN \cite{DBLP:journals/corr/Long015}, D-CORAL \cite{dcoral}, RTN \cite{DBLP:journals/corr/Long0J16} and JAN \cite{DBLP:journals/corr/Long0J16a} architecture where Alexnet \cite{NIPS2012_4824} is used, comprising of five convolution layers and three fully connected layers.

\subsection{Experimental Setup}

For synthetic image generation, we use ComboGAN \cite{DBLP:journals/corr/abs-1712-06909} where we set the value of $\lambda$ as $10$ in Equation \ref{eq:3}. In this model, Adam solver is used with a batch size of 1. We trained all the networks from scratch and we set the learning rate to 0.0002. We set total 200 epochs, and for the first 100 epochs, the learning rate does not change whereas it linearly decays the rate towards zero over the next 100 epochs. 

For domain generalization, we use two streams of CNN. In each stream, we extend AlexNet \cite{NIPS2012_4824} model which is pretrained on the ImageNet \cite{5206848} dataset. We set the dimension of the last fully connected layer (fc8) to the number of categories (for instance, 65 for Office-Home dataset). We set the learning rate to 0.0001 to optimize the network. Moreover, we set the batch size to 128, weight decay to $5 \times 10^{-4}$ and momentum to 0.9 during the training phase. 

\subsection{Results and Discussion}

In this section, we will discuss our experimental results in detail. For fair comparison, we use the same network architecture (AlexNet \cite{NIPS2012_4824}) that are used in the existing domain adaptation methods such as DAN \cite{DBLP:journals/corr/Long015}, D-CORAL \cite{dcoral}, JAN \cite{DBLP:journals/corr/Long0J16a} and RTN \cite{DBLP:journals/corr/Long0J16} in domain generalization settings. DAN \cite{DBLP:journals/corr/Long015} is a deep domain adaptation model where the discrepancy between the source and target data is minimized using MMD. JAN \cite{DBLP:journals/corr/Long0J16a} is also a deep domain adaptation method where the discrepancy between the source and target data is mitigated using Joint Maximum Mean Discrepancy (JMMD) criterion. RTN \cite{DBLP:journals/corr/Long0J16} is used for minimizing domain discrepancy between different distributions data using residual transfer network and MMD. On the other hand, D-CORAL \cite{dcoral} is a DDA method where the discrepancy between two domains is minimized by the second order statistics (covariances) which is known as correlation alignment.

%\restylefloat{table}
\begin{table}[!htbp]
\fontsize{6}{7.0}\selectfont 
\begin{center}
%\begin{tabular}{|l||c|c|c|c|c|c||cc}

\begin{tabular}{|p{1.8cm}|p{0.70cm}|p{1.0cm}|p{0.70cm}|p{0.70cm}|}

%{l||c|c|c|c|c|c|c|c||cc}
%\rowcolor[gray]{0.95}
\hline

%\textbf{Sources $\rightarrow$ Target} &uDICA &LRE-SVM &MTAE & DSN & DBADG & DAN & D-CORAL & JAN & RTN \\

\textbf{Sources $\rightarrow$ Target}  & \textbf{Ours ($DAN$)} & \textbf{Ours ($D-CORAL$)} & \textbf{Ours ($JAN$)} & \textbf{Ours ($RTN$)}  \\

\hline\hline
\textbf{$Pr, Rw, Cl \rightarrow $Ar} &44.16  &44.08 & \textbf{ 45.29} & 44.58    \\
\hline

\textbf{$Pr, Rw, Ar \rightarrow $Cl} &39.99 &40.28   &\textbf{40.70} & 40.61  \\
\hline

\textbf{$Pr, Ar, Cl \rightarrow $Rw} &64.17 &64.10 &\textbf{ 64.89}& 64.33    \\
\hline

 \textbf{$Ar, Cl, Rw \rightarrow $Pr} &61.99 &62.82 &\textbf{63.78} & 63.05 \\
\hline

\hline
\textbf{$Avg.$} &52.58 &52.82 &\textbf{53.67}&53.14 \\
\hline

\end{tabular}
\end{center}
\caption{Recognition accuracies for domain generalization on the Office-Home dataset \cite{venkateswara2017Deep}. Here, we split the source datasets into 70-30\% training-validation samples.}
\label{table_off-home}
\end{table}

%%%%%%%%%%%%%%%%%%%%
%---------------------------------COMBOGAN
%\restylefloat{table}
\begin{table}[!htbp]
\fontsize{6}{7.0}\selectfont 
\begin{center}
%\begin{tabular}{|l||c|c|c|c|c|c}
%{l||c|c|c|c|c|c|c|c||cc}
%\rowcolor[gray]{0.95}

\begin{tabular}{|p{1.8cm}|p{0.70cm}|p{1.0cm}|p{0.70cm}|p{0.70cm}|}

\hline

%\textbf{Sources $\rightarrow$ Target} &uDICA &LRE-SVM &MTAE & DSN & DBADG & DAN & D-CORAL & JAN & RTN \\

\textbf{Sources $\rightarrow$ Target}  & \textbf{Ours ($DAN_S$)} & \textbf{Ours ($D-CORAL_S$)} & \textbf{Ours ($JAN_S$)} & \textbf{Ours ($RTN_S$)}  \\

\hline\hline
\textbf{$Pr, Rw, Cl \rightarrow $Ar} &47.85  &47.98 &\textbf{48.09} &  47.90  \\
\hline

\textbf{$Pr, Rw, Ar \rightarrow $Cl} &43.85 &44.73   &45.20 &  \textbf{45.30} \\
\hline

\textbf{$Pr, Ar, Cl \rightarrow $Rw} &67.27 &67.58 &\textbf{68.35} & 68.09    \\
\hline

 \textbf{$Ar, Cl, Rw \rightarrow $Pr} &64.29 &64.78 &\textbf{66.52} & 66.21 \\
\hline

\textbf{$Avg.$} &55.82&56.27&\textbf{57.04}&56.88 \\
\hline

\end{tabular}
\end{center}
\caption{Recognition accuracies for domain generalization on the Office-Home dataset \cite{venkateswara2017Deep} with synthetic images that are generated using ComboGAN. The subscript \textbf{S} represents synthetic data.}
\label{table_off-home_DG}
\end{table}

%---------------------------------MUNIT
%%%%%%%%%%%%%%%%%%%%

%\restylefloat{table}
\begin{table}[!htbp]
\fontsize{6}{7.0}\selectfont 
\begin{center}
%\begin{tabular}{|l||c|c|c|c|c|c}

\begin{tabular}{|p{1.8cm}|p{0.70cm}|p{1.0cm}|p{0.70cm}|p{0.70cm}|}

%{l||c|c|c|c|c|c|c|c||cc}
%\rowcolor[gray]{0.95}
\hline

%\textbf{Sources $\rightarrow$ Target} &uDICA &LRE-SVM &MTAE & DSN & DBADG & DAN & D-CORAL & JAN & RTN \\

\textbf{Sources $\rightarrow$ Target}  & \textbf{Ours ($DAN_S$)} & \textbf{Ours ($D-CORAL_S$)} & \textbf{Ours ($JAN_S$)} & \textbf{Ours ($RTN_S$)}  \\

\hline\hline
\textbf{$Pr, Rw, Cl \rightarrow $Ar} &45.03  &45.27 &\textbf{46.24} &  45.71  \\
\hline

\textbf{$Pr, Rw, Ar \rightarrow $Cl} &41.92 &42.60   &\textbf{42.89} & 42.85 \\
\hline

\textbf{$Pr, Ar, Cl \rightarrow $Rw} &65.39 &65.72 &\textbf{66.08} & 65.41    \\
\hline

 \textbf{$Ar, Cl, Rw \rightarrow $Pr} &62.56 &63.43 &\textbf{64.90} & 64.37 \\
\hline

\textbf{$Avg.$} &53.73 &54.26 &\textbf{54.78}&54.59 \\
\hline

\end{tabular}
\end{center}
\caption{Recognition accuracies for domain generalization on the Office-Home dataset \cite{venkateswara2017Deep} with synthetic images that are generated using MUNIT. The subscript \textbf{S} represents synthetic data.}
\label{table_off-home_DG_unit}
\end{table}

%---------------------------------Stargan
%%%%%%%%%%%%%%%%%%%%

%\restylefloat{table}
\begin{table}[!htbp]
\fontsize{6}{7.0}\selectfont 
\begin{center}
%\begin{tabular}{|l||c|c|c|c|c|c}

\begin{tabular}{|p{1.8cm}|p{0.70cm}|p{1.0cm}|p{0.70cm}|p{0.70cm}|}

%{l||c|c|c|c|c|c|c|c||cc}
%\rowcolor[gray]{0.95}
\hline

%\textbf{Sources $\rightarrow$ Target} &uDICA &LRE-SVM &MTAE & DSN & DBADG & DAN & D-CORAL & JAN & RTN \\

\textbf{Sources $\rightarrow$ Target}  & \textbf{Ours ($DAN_S$)} & \textbf{Ours ($D-CORAL_S$)} & \textbf{Ours ($JAN_S$)} & \textbf{Ours ($RTN_S$)}  \\

\hline\hline
\textbf{$Pr, Rw, Cl \rightarrow $Ar} &44.86  &45.09 &\textbf{45.95} &  45.26  \\
\hline

\textbf{$Pr, Rw, Ar \rightarrow $Cl} &40.57 &42.19   &42.38 & \textbf{42.41} \\
\hline

\textbf{$Pr, Ar, Cl \rightarrow $Rw} &64.80 &65.38 &\textbf{65.94} & 65.05    \\
\hline

 \textbf{$Ar, Cl, Rw \rightarrow $Pr} &62.07 &63.10 &\textbf{64.14} & 63.51 \\
\hline

\textbf{$Avg.$} &53.08 &53.69 &\textbf{54.60}&54.06 \\
\hline

\end{tabular}
\end{center}
\caption{Recognition accuracies for domain generalization on the Office-Home dataset \cite{venkateswara2017Deep} with synthetic images that are generated using Stargan. The subscript \textbf{S} represents synthetic data.}
\label{table_off-home_DG_stargan}
\end{table}
%---------------------------------

%---------------------------------

%%%%%%%%%%%%%%%%%%%

%\restylefloat{table}
\begin{table*}[!htbp]
\fontsize{6}{7.0}\selectfont 
\begin{center}
\begin{tabular}{|l||c|c|c|c|c|c|c|c|c|c|c|cc}
%\rowcolor[gray]{0.95}
\hline

\textbf{Sources $\rightarrow$ Target} &Undo-Bias&UML &LRE-SVM & MTAE& DSN& DGLRC & \textbf{Ours ($DAN_S$)} & \textbf{Ours ($D-CORAL_S$)} & \textbf{Ours ($JAN_S$)} & \textbf{Ours ($RTN_S$)} \\
\hline\hline
\textbf{$A, W \rightarrow $D} &98.45 &98.76 &98.84 &98.97 &99.02 &\textbf{99.44} &97.18 &96.18 &97.87 &97.23 \\
\hline

\textbf{$A, D \rightarrow $W} &93.38&93.76&93.98&94.21 &94.45  &\textbf{95.28} &92.70&93.20 &94.03 &94.53 \\
\hline

\textbf{$D, W \rightarrow $A}&42.43 &41.65 &44.14&43.67 &43.98 &45.36 &49.30 &51.61 &\textbf{51.73} &50.86 \\
\hline
\textbf{Avg.} &78.08&78.05&78.99&78.95&79.15&80.02&79.72&80.33&\textbf{81.21}&80.87 \\
\hline

\end{tabular}
\end{center}
\caption{Recognition accuracies for domain generalization on the Office31 dataset \cite{Saenko:2010:AVC:1888089.1888106} using synthetic images that are generated by ComboGAN. The subscript \textbf{S} represents synthetic data.}
\label{table_office31}
\end{table*}

%\restylefloat{table}
\begin{table*}[!htbp]
\fontsize{6}{7.0}\selectfont 
\begin{center}
\begin{tabular}{|l||c|c|c|c|c|c|c|c|c|c|c|c|c|c}
%\rowcolor[gray]{0.95}
\hline

\textbf{Sources $\rightarrow$ Target} &Undo-Bias & UML &LRE-SVM &MTAE &DSN & DGLRC  & \textbf{Ours ($DAN_S$)} & \textbf{Ours ($D-CORAL_S$)} & \textbf{Ours ($JAN_S$)} & \textbf{Ours ($RTN_S$)}  \\
\hline\hline
\textbf{$W, D, C \rightarrow $A} &90.98&91.02 &91.87&93.13&- &\textbf{94.21}&92.27 &92.79 &93.31 &93.06  \\
\hline

\textbf{$A, W, D \rightarrow $C} &85.95 &84.59 &86.38 &86.15&-&\textbf{87.63} & 84.41& 86.74 &86.28 &85.31 \\
\hline

\textbf{$A, C \rightarrow $D, W} &80.49  &82.29 &84.59 &85.35 &85.76&\textbf{86.32}& 85.17&82.07 & 85.17 &84.27  \\
\hline

\textbf{$D, W \rightarrow $A, C} &69.98  &79.54 &81.17 &80.52&81.22&82.24 &80.24 &79.56 &\textbf{82.74} &81.53\\
\hline 
\textbf{$Avg.$} &81.85 &84.36 &86.00 &86.28 &- &\textbf{87.60} &85.52 &85.29 &86.87 &86.04\\
\hline 

\end{tabular}
\end{center}
\caption{Recognition accuracies for domain generalization on the Office-Caltech dataset \cite{DBLP:conf/cvpr/GongSSG12} using synthetic images that are generated by ComboGAN. The subscript \textbf{S} represents synthetic data.}
\label{table_offcal}
\end{table*}

\restylefloat{table}
\begin{table*}[!htbp]
\fontsize{6}{7.0}\selectfont 
\begin{center}
\begin{tabular}{|l||c|c|c|c|c|c|c|c|c|c|c}
%\rowcolor[gray]{0.95}
\hline

\textbf{Sources $\rightarrow$ Target} &uDICA &LRE-SVM &MTAE & DSN & DBADG & \textbf{Ours ($DAN_S$)} & \textbf{Ours ($D-CORAL_S$)} & \textbf{Ours ($JAN_S$)} & \textbf{Ours ($RTN_S$)}  \\
\hline\hline
\textbf{$P, C, S \rightarrow $A} &64.57 &59.74 &60.27 &61.13 &62.86 &\textbf{64.69} &61.37 &62.64 &62.64   \\
\hline

\textbf{$P, S, A \rightarrow $C} &64.54 &52.81 &58.65 &66.54  &\textbf{66.97} & 63.60&66.16 &65.98 &66.52 \\
\hline

\textbf{$S, A, C \rightarrow $P} &\textbf{91.78} &85.53 &91.12 &83.25  &89.50 &90.11 &89.16 & 90.44 &89.86  \\
\hline

 \textbf{$P, C, A \rightarrow $S} &51.12 &37.89&47.86 &58.58 &57.51 &58.08 &\textbf{58.92} &58.76 & 57.68\\
\hline

\textbf{$Avg.$} &68.00&58.99&64.48&63.38&69.21&69.12&68.90&\textbf{69.45} &69.18 \\
\hline

\end{tabular}
\end{center}
\caption{Recognition accuracies for domain generalization on the PACS dataset \cite{8237853} using synthetic images that are generated by ComboGAN. The subscript \textbf{S} represents synthetic data.}
\label{table_pacs}
\end{table*}

We evaluate our DG model on different datasets. At first, we follow the same experimental setting as in \cite{DBLP:journals/pami/GhifaryBKZ17,DBLP:conf/iccv/GhifaryKZB15} for Office-Home dataset, and randomly split each source domain into a training set ($70\%$ ) and a validation set ($30\%$). Then we conduct experiments of having one domain totally unseen during the training stage. The $70\%$ data is fed into one stream of CNN and $30\%$ data is fed into the second stream of CNN. In the test phase, we use unlabeled target data which is completely unseen during the training stage. It is noted that the target data is not splitting. In the above setting, we evaluate 4 existing domain adaptation (DAN \cite{DBLP:journals/corr/Long015}, D-CORAL \cite{dcoral}, JAN \cite{DBLP:journals/corr/Long0J16a} and RTN \cite{DBLP:journals/corr/Long0J16}) methods for domain generalization settings. We report these comparative results in Table 1. 

After that, we generate synthetic images using ComboGAN \cite{DBLP:journals/corr/abs-1712-06909} in the training phase. Figure \ref{fig:office} shows some generated images. The real images are fed into one stream of CNN and synthetic images are fed into another stream of CNN. In Table 2, we report the comparative results. From Table 1 and Table 2, we can observe that our method improves on an average 3\% higher than 70\%-30\% settings in each transfer task on Office-Home dataset.  

To further increase the justification of adopting ComboGAN in our framework, we also experimentally evaluate our domain generalization model using synthetic images that are generated by MUNIT\cite{MUNIT} and Stargan \cite{starGAN} on Office-Home dataset. The domain generalization on different tasks are reported in Table 3 and Table 4 using MUNIT and Stargan respectively. From the results (Tables 1, 2, 3 and 4), we make two important observations: (1) The multi-component synthetic data generation using GAN boosts the domain generalization performance; and (2) As ComboGAN can handle more than two domains at a time compared to MUNIT and Stargan, it can generate more multi-component images which are more effective for domain generalization.

We further evaluate and compare our proposed approach
with both shallow and deep domain generalization  state-of-the-art methods: Undoing the Damage of Dataset Bias (Undo-Bias) \cite{Khosla:2012:UDD:2402940.2402953}, Unbiased Metric Learning (UML) \cite{6751316}, Low-Rank Structure from Latent Domains for Domain Generalization (LRE-SVM) \cite{10.1007/978-3-319-10578-9_41}, Multi-Task Autoencoders (MTAE) \cite{DBLP:conf/iccv/GhifaryKZB15}, Domain Separation Network (DSN) \cite{Bousmalis:2016:DSN:3157096.3157135}, Deep Domain Generalization with Structured
Low-Rank Constraint(DGLRC) \cite{8053784}, Domain Generalization via Invariant Feature Representation (uDICA) \cite{Muandet:2013:DGV:3042817.3042820}, Deeper, Broader and Artier Domain Generalization (DBADG) \cite{8237853}.

We report comparative results in Table 5, 6 and 7 on Office 31, Office-Caltech, and PACS datasets respectively. From Table \ref{table_office31}, it can be seen that, the previous best method \cite{8053784} achieved 80.02\% average accuracy where the domain generalization issue is solved by using a structured low rank constraint. In contrast, our domain generalization method using synthetic data with D-CORAL \cite{dcoral}, JAN \cite{DBLP:journals/corr/Long0J16a}, and RTN \cite{DBLP:journals/corr/Long0J16} network architectures achieves 80.33\%, 81.21\% and 80.87\% average accuracies respectively which outperforms the state-of-the-art methods. For Office-Caltech dataset (see Table \ref{table_offcal}), although DGLRC  \cite{8053784} achieved best performance, our proposed method is different in network architecture as we use generative adversarial network to generate synthetic data. Using synthetic data with DAN \cite{DBLP:journals/corr/Long015}, D-CORAL \cite{dcoral}, JAN \cite{DBLP:journals/corr/Long0J16a} and RTN \cite{DBLP:journals/corr/Long0J16} network architecture, we achieve 85.52\%, 85.29\%, 86.87\% and 86.04\% average accuracies respectively. For PACS dataset, our proposed method achieves state-of-the-art performance using synthetic data and JAN architecture \cite{DBLP:journals/corr/Long0J16a}. It is worth noting that for PACS dataset, our result is 69.45\% while the previous state-of-the-art method achieved 69.21\%.

% Our model is capable to translate images from one domain to other domains even when the domains do not share the same class labels. Our model takes the overall information from the domains, it translate images from one domain to other domains whose styles are similar to the corresponding domains. On the other hand, for minimizing the discrepancy among these domains, we take account close set domain generalization settings where we assume that every source domains share the same class labels. In our future work, we will explore our method on open-set scene where the source domains may not share the same class labels and source domains may have different classes than target domain.

Our image translation model is unsupervised, and capable of translating images from one domain to another in open-set domain adaptation/generalization scenario. However, to minimize discrepancy among different domains, we considered close set domain generalization settings where we assume that every source domains share similar classes as the target domain. In the future work, we will explore our method on outdoor scene dataset where the source domains may not share the same class labels with the target domain.

\section{Conclusion}
In this paper, we developed a novel deep domain generalization architecture using synthetic images which were generated by a GAN and existing domain discrepancy minimizing metrics which aims to learn a domain agnostic model from the real and synthetic data that can be applied for unseen datasets. More specifically, we built an architecture to transfer the style from one domain image to another domain. After generating synthetic data, we used maximum mean discrepancy and correlation alignment metrics to minimize the discrepancy between the synthetic data and real data. Extensive experimental results on several benchmark datasets demonstrate our proposed method achieves state-of-the-art performance.

\section*{Acknowledgement}
The research presented in this paper was supported by Australian Research Council (ARC) Discovery Project
Grants DP170100632 and DP140100793.

{\small
\bibliographystyle{ieee}
\bibliography{paper}

\begin{thebibliography}{10}\itemsep=-1pt

\bibitem{DBLP:journals/corr/abs-1712-06909}
A.~Anoosheh, E.~Agustsson, R.~Timofte, and L.~V. Gool.
\newblock Combogan: Unrestrained scalability for image domain translation.
\newblock {\em CoRR}, abs/1712.06909, 2017.

\bibitem{NIPS2011_4312}
G.~Blanchard, G.~Lee, and C.~Scott.
\newblock Generalizing from several related classification tasks to a new
  unlabeled sample.
\newblock In {\em Advances in Neural Information Processing Systems (NIPS)}.
  2011.

\bibitem{Bousmalis:2016:DSN:3157096.3157135}
K.~Bousmalis, G.~Trigeorgis, N.~Silberman, D.~Krishnan, and D.~Erhan.
\newblock Domain separation networks.
\newblock In {\em Proceedings of the International Conference on Neural
  Information Processing Systems (NIPS)}, 2016.

\bibitem{starGAN}
Y.~Choi, M.~Choi, M.~Kim, J.~Ha, S.~Kim, and J.~Choo.
\newblock Stargan: Unified generative adversarial networks for multi-domain
  image-to-image translation.
\newblock In {\em Conference on Computer Vision and Pattern Recognition
  (CVPR)}, 2018.

\bibitem{DBLP:series/acvpr/978-3-319-58347-1}
G.~Csurka.
\newblock {\em Domain Adaptation in Computer Vision Applications}.
\newblock Advances in Computer Vision and Pattern Recognition. Springer, 2017.

\bibitem{5206848}
J.~Deng, W.~Dong, R.~Socher, L.~J. Li, K.~Li, and L.~Fei-Fei.
\newblock Imagenet: A large-scale hierarchical image database.
\newblock In {\em Computer Vision and Pattern Recognition (CVPR)}, 2009.

\bibitem{8053784}
Z.~Ding and Y.~Fu.
\newblock Deep domain generalization with structured low-rank constraint.
\newblock {\em IEEE Transactions on Image Processing}, 27(1):304--313, 2018.

\bibitem{Donahue:2014:DDC:3044805.3044879}
J.~Donahue, Y.~Jia, O.~Vinyals, J.~Hoffman, N.~Zhang, E.~Tzeng, and T.~Darrell.
\newblock Decaf: A deep convolutional activation feature for generic visual
  recognition.
\newblock In {\em International Conference on Machine Learning (ICML)}, 2014.

\bibitem{6751316}
C.~Fang, Y.~Xu, and D.~N. Rockmore.
\newblock Unbiased metric learning: On the utilization of multiple datasets and
  web images for softening bias.
\newblock In {\em IEEE International Conference on Computer Vision (ICCV)},
  2013.

\bibitem{Ganin:2016:DTN:2946645.2946704}
Y.~Ganin, E.~Ustinova, H.~Ajakan, P.~Germain, H.~Larochelle, F.~Laviolette,
  M.~Marchand, and V.~Lempitsky.
\newblock Domain-adversarial training of neural networks.
\newblock {\em Journal of Machine Learning Research}, 17(1):2096--2030, 2016.

\bibitem{DBLP:journals/pami/GhifaryBKZ17}
M.~Ghifary, D.~Balduzzi, W.~B. Kleijn, and M.~Zhang.
\newblock Scatter component analysis: {A} unified framework for domain
  adaptation and domain generalization.
\newblock {\em {IEEE} Trans. Pattern Anal. Mach. Intell.}, 39(7):1414--1430,
  2017.

\bibitem{DBLP:conf/iccv/GhifaryKZB15}
M.~Ghifary, W.~B. Kleijn, M.~Zhang, and D.~Balduzzi.
\newblock Domain generalization for object recognition with multi-task
  autoencoders.
\newblock In {\em IEEE International Conference on Computer Vision, (ICCV)},
  2015.

\bibitem{DBLP:conf/cvpr/GongSSG12}
B.~Gong, Y.~Shi, F.~Sha, and K.~Grauman.
\newblock Geodesic flow kernel for unsupervised domain adaptation.
\newblock In {\em IEEE Conference on Computer Vision and Pattern Recognition
  (CVPR)}, 2012.

\bibitem{NIPS2014_5423}
I.~Goodfellow, J.~Pouget-Abadie, M.~Mirza, B.~Xu, D.~Warde-Farley, S.~Ozair,
  A.~Courville, and Y.~Bengio.
\newblock Generative adversarial nets.
\newblock In {\em Advances in Neural Information Processing Systems (NIPS)}.
  2014.

\bibitem{griffinHolubPerona}
G.~Griffin, A.~Holub, and P.~Perona.
\newblock Caltech-256 object category dataset.
\newblock Technical report, California Institute of Technology, 2007.

\bibitem{MUNIT}
X.~Huang, M.-Y. Liu, S.~Belongie, and J.~Kautz.
\newblock Multimodal unsupervised image-to-image translation.
\newblock In {\em European Conference on Computer Vision (ECCV)}, 2018.

\bibitem{DBLP:conf/cvpr/IsolaZZE17}
P.~Isola, J.~Zhu, T.~Zhou, and A.~A. Efros.
\newblock Image-to-image translation with conditional adversarial networks.
\newblock In {\em Conference on Computer Vision and Pattern Recognition
  (CVPR)}, 2017.

\bibitem{8354250}
A.~Khatun, S.~Denman, S.~Sridharan, and C.~Fookes.
\newblock A deep four-stream siamese convolutional neural network with joint
  verification and identification loss for person re-detection.
\newblock In {\em IEEE Winter Conference on Applications of Computer Vision
  (WACV)}, 2018.

\bibitem{Khosla:2012:UDD:2402940.2402953}
A.~Khosla, T.~Zhou, T.~Malisiewicz, A.~A. Efros, and A.~Torralba.
\newblock Undoing the damage of dataset bias.
\newblock In {\em Proceedings of the European Conference on Computer Vision
  (ECCV)}, 2012.

\bibitem{DBLP:journals/corr/KoniuszTP16}
P.~Koniusz, Y.~Tas, and F.~Porikli.
\newblock Domain adaptation by mixture of alignments of second-or higher-order
  scatter tensors.
\newblock In {\em IEEE Conference on Computer Vision and Pattern Recognition
  (CVPR)}, 2017.

\bibitem{NIPS2012_4824}
A.~Krizhevsky, I.~Sutskever, and G.~E. Hinton.
\newblock Imagenet classification with deep convolutional neural networks.
\newblock In {\em Neural Information Processing Systems (NIPS)}. 2012.

\bibitem{8237853}
D.~Li, Y.~Yang, Y.~Z. Song, and T.~M. Hospedales.
\newblock Deeper, broader and artier domain generalization.
\newblock In {\em IEEE International Conference on Computer Vision (ICCV)},
  2017.

\bibitem{DBLP:journals/corr/Long015}
M.~Long, Y.~Cao, J.~Wang, and M.~I. Jordan.
\newblock Learning transferable features with deep adaptation networks.
\newblock In {\em International Conference on Machine Learning (ICML)}, 2015.

\bibitem{DBLP:journals/corr/Long0J16}
M.~Long, H.~Zhu, J.~Wang, and M.~I. Jordan.
\newblock Unsupervised domain adaptation with residual transfer networks.
\newblock In {\em Conference on Neural Information Processing Systems (NIPS)},
  2016.

\bibitem{DBLP:journals/corr/Long0J16a}
M.~Long, H.~Zhu, J.~Wang, and M.~I. Jordan.
\newblock Deep transfer learning with joint adaptation networks.
\newblock In {\em International Conference on Machine Learning (ICML)}, 2017.

\bibitem{DBLP:journals/corr/MorerioM17}
P.~Morerio and V.~Murino.
\newblock Correlation alignment by riemannian metric for domain adaptation.
\newblock {\em CoRR}, abs/1705.08180, 2017.

\bibitem{motiian2017CCSA}
S.~Motiian, M.~Piccirilli, D.~A. Adjeroh, and G.~Doretto.
\newblock Unified deep supervised domain adaptation and generalization.
\newblock In {\em IEEE International Conference on Computer Vision (ICCV)},
  2017.

\bibitem{Muandet:2013:DGV:3042817.3042820}
K.~Muandet, D.~Balduzzi, and B.~Sch\"{o}lkopf.
\newblock Domain generalization via invariant feature representation.
\newblock In {\em Proceedings of the International Conference on International
  Conference on Machine Learning (ICML)}, 2013.

\bibitem{7410834}
L.~Niu, W.~Li, and D.~Xu.
\newblock Multi-view domain generalization for visual recognition.
\newblock In {\em IEEE International Conference on Computer Vision (ICCV)},
  2015.

\bibitem{7733141}
L.~Niu, W.~Li, D.~Xu, and J.~Cai.
\newblock An exemplar-based multi-view domain generalization framework for
  visual recognition.
\newblock {\em IEEE Transactions on Neural Networks and Learning Systems},
  29(2):259--272, 2018.

\bibitem{5640675}
S.~J. Pan, I.~W. Tsang, J.~T. Kwok, and Q.~Yang.
\newblock Domain adaptation via transfer component analysis.
\newblock {\em IEEE Transactions on Neural Networks}, 22, 2011.

\bibitem{7078994}
V.~M. Patel, R.~Gopalan, R.~Li, and R.~Chellappa.
\newblock Visual domain adaptation: A survey of recent advances.
\newblock {\em IEEE Signal Processing Magazine}, 32(3):53--69, 2015.

\bibitem{Perarnau2016}
G.~Perarnau, J.~van~de Weijer, B.~Raducanu, and J.~M. \'Alvarez.
\newblock {Invertible Conditional GANs for image editing}.
\newblock In {\em Neural Information Processing Systems (NIPS) Workshop on
  Adversarial Training}, 2016.

\bibitem{Reed:2016:GAT:3045390.3045503}
S.~Reed, Z.~Akata, X.~Yan, L.~Logeswaran, B.~Schiele, and H.~Lee.
\newblock Generative adversarial text to image synthesis.
\newblock In {\em International Conference on Machine Learning (ICML)}, 2016.

\bibitem{Russell2008}
B.~C. Russell, A.~Torralba, K.~P. Murphy, and W.~T. Freeman.
\newblock Labelme: A database and web-based tool for image annotation.
\newblock {\em International Journal of Computer Vision}, 2008.

\bibitem{Saenko:2010:AVC:1888089.1888106}
K.~Saenko, B.~Kulis, M.~Fritz, and T.~Darrell.
\newblock Adapting visual category models to new domains.
\newblock In {\em European Conference on Computer Vision (ECCV)}, 2010.

\bibitem{NIPS2007_3248}
M.~Sugiyama, S.~Nakajima, H.~Kashima, P.~V. Buenau, and M.~Kawanabe.
\newblock Direct importance estimation with model selection and its application
  to covariate shift adaptation.
\newblock In {\em Advances in Neural Information Processing Systems (NIPS)}.
  2008.

\bibitem{coral}
B.~Sun, J.~Feng, and K.~Saenko.
\newblock Return of frustratingly easy domain adaptation.
\newblock In {\em Association for the Advancement of Artificial Intelligence
  (AAAI)}, 2016.

\bibitem{dcoral}
B.~Sun and K.~Saenko.
\newblock Deep coral: Correlation alignment for deep domain adaptation.
\newblock In {\em European Conference on Computer Vision (ECCV) Workshops},
  2016.

\bibitem{DBLP:journals/corr/TzengHDS15}
E.~Tzeng, J.~Hoffman, T.~Darrell, and K.~Saenko.
\newblock Simultaneous deep transfer across domains and tasks.
\newblock In {\em International Conference on Computer Vision (ICCV)}, 2015.

\bibitem{DBLP:journals/corr/TzengHZSD14}
E.~Tzeng, J.~Hoffman, N.~Zhang, K.~Saenko, and T.~Darrell.
\newblock Deep domain confusion: Maximizing for domain invariance.
\newblock {\em CoRR}, abs/1412.3474, 2014.

\bibitem{venkateswara2017Deep}
H.~Venkateswara, J.~Eusebio, S.~Chakraborty, and S.~Panchanathan.
\newblock Deep hashing network for unsupervised domain adaptation.
\newblock In {\em Conference on Computer Vision and Pattern Recognition
  (CVPR)}, 2017.

\bibitem{10.1007/978-3-319-10578-9_41}
Z.~Xu, W.~Li, L.~Niu, and D.~Xu.
\newblock Exploiting low-rank structure from latent domains for domain
  generalization.
\newblock In {\em European Conference on Computer Vision (ECCV)}, 2014.

\bibitem{Zadrozny:2004:LEC:1015330.1015425}
B.~Zadrozny.
\newblock Learning and evaluating classifiers under sample selection bias.
\newblock In {\em International Conference on Machine Learning (ICML)}, 2004.

\bibitem{CycleGAN2017}
J.~Y. Zhu, T.~Park, P.~Isola, and A.~A. Efros.
\newblock Unpaired image-to-image translation using cycle-consistent
  adversarial networks.
\newblock In {\em IEEE International Conference on Computer Vision (ICCV)},
  2017.

\bibitem{bicycleGAN}
J.-Y. Zhu, R.~Zhang, D.~Pathak, T.~Darrell, A.~A. Efros, O.~Wang, and
  E.~Shechtman.
\newblock Toward multimodal image-to-image translation.
\newblock In {\em Advances in Neural Information Processing Systems (NIPS)}.
  2017.

\end{thebibliography}
}

\end{document}